\documentclass[10pt,twocolumn,letterpaper]{article}

\usepackage{cvpr}              

\usepackage[table]{xcolor}

\usepackage{algorithm}
\usepackage{algpseudocode}
\definecolor{cvprblue}{rgb}{0.21,0.49,0.74}
\usepackage[pagebackref,breaklinks,colorlinks,allcolors=cvprblue]{hyperref}

\def\paperID{10139} 
\def\confName{CVPR}
\def\confYear{2026}

\title{Beyond Deepfake vs Real: Facial Deepfake Detection in the Open-Set Paradigm}

\author{
    Thiru Thillai Nadarasar Bahavan$^1$, 
    Sachith Seneviratne$^1$, 
    Sanjay Saha$^2$, \\
    Ken Chen$^1$, 
    Sanka Rasnayaka$^2$, 
    Saman Halgamuge$^1$ \\
    $^1$The University of Melbourne, Parkville \quad $^2$National University of Singapore \\
    {\tt\small \{bahavant, ken.chen2\}@student.unimelb.edu.au, \{sachith.seneviratne, saman\}@unimelb.edu.au} \\
    {\tt\small \{contact.sanjaysaha, sanka\}@nus.edu.sg}
}

\begin{document}
\maketitle
\begin{abstract}

Facial forgery methods such as deepfakes can be misused for identity manipulation and spreading misinformation. They have evolved alongside advancements in generative AI, leading to new and more sophisticated forgery techniques that diverge from existing ``known" methods. Conventional deepfake detection methods use the closed-set paradigm, thus limiting their applicability to detecting forgeries created using methods that are not part of the training dataset. In this paper, we propose a shift from the closed-set paradigm for deepfake detection. In the open-set paradigm, models are designed not only to identify images created by known facial forgery methods but also to identify and flag those produced by previously unknown methods as ``unknown" and not as unforged/real/unmanipulated. In this paper, we propose an open-set deepfake classification algorithm based on supervised contrastive learning. The open-set paradigm used in our model allows it to function as a more robust tool capable of handling emerging and unseen deepfake techniques, enhancing reliability and confidence, and complementing forensic analysis. In the open-set paradigm, we identify three groups, including the ``unknown” group that is neither considered a known deepfake nor real. We investigate deepfake open-set classification across three scenarios: classifying deepfakes from unknown methods not as real,  distinguishing real images from deepfakes, and classifying deepfakes from known methods, using the FaceForensics++ dataset as a benchmark. Our method achieves state-of-the-art results in the first two tasks and competitive results in the third task. 
\end{abstract}

\section{Introduction}

\begin{figure}[htbp]
    \centering
    \includegraphics[width=\linewidth]{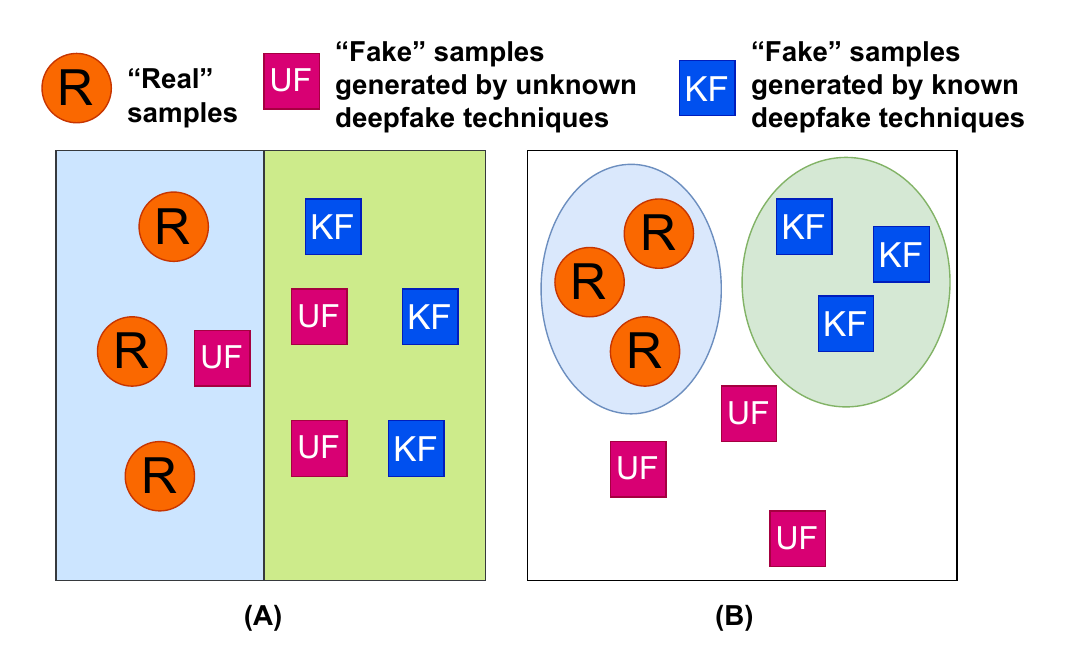}
    \caption{
        This figure illustrates the latent space and decision boundaries of both the closed-set and open-set paradigms. (A) In the case of a closed-set classifier, the model learns a feature space and decision boundaries based solely on the known deepfake techniques and real data, distinguishing between Real and Known Deepfakes. (B) However, this approach cannot identify samples from previously unseen techniques. In contrast, open-set recognition establishes more flexible decision boundaries around the known classes and learns more general features.
    }
    \label{fig:mainidea}
\end{figure}

Facial deepfake methods are deep learning-based algorithms that generate synthetic media, such as videos or images, by convincingly mimicking real people's faces.  In this context, real or unmanipulated refers to authentic, unaltered data, often described as pristine, while forged or manipulated refers to media altered through deepfake techniques. Facial deepfakes have gained widespread notoriety due to their capacity to produce strikingly realistic visual content, which can be misused by malicious entities to spread false information, manipulate identities, and infringe upon personal privacy \cite{dhs2023increasing,kpmg2023deepfakes}. The recent advancement of generative AI, coupled with increased GPU capacity and availability, has led to new techniques for creating facial deepfakes.

Most existing deepfake detectors are built on two assumptions: The testing dataset is created using the same forgery techniques used to make the training dataset, where both training and testing data share the same labels and feature space, and both datasets follow the same distribution. However, in real-world scenarios, it is difficult to collect training data that covers all possible test categories. As a result, an image from an unknown forgery technique—a technique not encountered during training—that violates these assumptions can be incorrectly identified as real (as shown in Fig.~\ref{fig:mainidea} (A)). Researchers have attempted to circumvent this problem by adding various inductive biases during the model design itself based on heuristics \cite{1swe,2swe,3swe,Rossler2019FaceForensics,qian2020thinking,38_paper}. However, these heuristics are derived from the observations of existing known deepfake techniques and may not apply to newer, emerging techniques. Given these implications, there is an urgent need to develop a reliable deepfake detection system to detect new and emerging variants of deepfake forgery methods. 

To overcome this limitation, open-set recognition (OSR) offers a more practical solution, requiring the model to not only achieve high accuracy on known classes but also to detect and manage samples from previously unseen categories during testing (as shown in Fig.~\ref{fig:mainidea} (B)). Open-set detection enables a model to recognize and appropriately handle data from classes that were not present in the training set. In the context of deepfake identification, this involves detecting not only known forgery methods but also identifying new, previously unseen forgery techniques. Rather than misclassifying these unknown forgeries as belonging to known classes, the model can flag them as unknown, thus maintaining robustness against emerging threats \cite{bendale2016towards}. 

This ability leads to several advantages: (A) It enhances robustness by helping models identify new deepfake methods, potentially reducing the risk of misclassification. (B) Early identification of emerging unknown forgeries enables the development of methods to prevent their future misuse by malicious actors. Forensic Experts can use few-shot or continual learning to quickly adapt models with limited data, ensuring robust detection of new deepfake techniques \cite{anonymous2023towards}.

Open-set methods operate by identifying the lack of expected features needed to classify a test sample as a known category, rather than detecting novel elements in the image, as suggested by the familiarity hypothesis \cite{famihyp}. This makes it crucial to develop highly discriminative features across manipulated and unmanipulated data. Supervised contrastive learning (SupCon) has proven effective for learning such discriminative features, with its success demonstrated in open-set recognition tasks \cite{9511382,xu2023contrastive}. However, adapting Supervised contrastive learning for deepfake detection introduces unique challenges.

In supervised contrastive learning, deepfake forgeries often form discriminative, well-separated representations, while real data tends to have more scattered representations. This makes it challenging to distinguish real from fake data. To address this, we modify the contrastive loss to emphasize the real class, encouraging tighter clustering of real data and making its representations more compact and distinct from manipulated data. This improvement enhances the model’s ability to detect deepfakes, particularly in open-set scenarios where unknown manipulations are identified as anomalies.

We introduce a method for open-set deepfake detection designed to address situations where the model encounters unknown forgery techniques. To the best of our knowledge, this is the first work to investigate the open-set classification problem in the context of deepfakes.

We make the following contributions.
\begin{itemize}
\item \textbf{Representation Learning algorithm}: We propose a variant of Supervised Contrastive Learning designed specifically for deepfake problems. 
\item \textbf{Good Performance in Unknown Deepfake Classification}: The proposed algorithm demonstrates excellent performance compared to the Xception baseline in detecting unknown deepfakes
\item \textbf{Competitive Performance in Known Deepfake Classification}: In the evaluation of known deepfakes, the proposed model performs competitively with state-of-the-art methods.

\item \textbf{Good Performance as an open-set detector}:
In distinguishing between real and deepfake data, we demonstrate that our feature learning method effectively captures a generalizable latent space. Our approach exhibits superior open-set performance within the cross-manipulation protocol, highlighting its robustness and efficacy in handling diverse manipulation techniques.

\end{itemize}

\section{Related Work}
\subsection{Facial Forgery/Deepfake Generation Methods}
Deepfake generation involves three primary approaches: Identity swapping, face-reenactment, and entire image synthesis. Identity swapping swaps the identity of faces in images or videos using techniques like auto-encoder-based methods (e.g., DeepFakes \cite{deepfakes2020}) and graphics-based swapping methods (e.g., FaceSwap \cite{kowalski2021faceswap}). Face-reenactment transfers facial expressions from a source video to a target video, maintaining the target person's identity while mimicking the source's expressions, with Face2Face \cite{thies2016face2face} and Neural Textures \cite{thies2019deferred} being prominent examples. Entire image synthesis generates completely new facial images without face-swapping, utilizing advanced generative models such as Generative Adversarial Networks (GANs) \cite{goodfellow2014generative}, diffusion models \cite{choi2018stargan,karras2019style,ho2020denoising}. These methods enable the creation of highly realistic synthetic media, each with unique capabilities and challenges.

\subsection{Generalizable Deepfake detectors} 
Deepfake detection faces significant challenges with generalization. Most of them are designed with a closed-set classifier. Current efforts are divided into image forgery and video forgery detection. In the realm of image forgery, innovative solutions have emerged, including frequency clues \cite{qian2020thinking,luo2021generalizing,liu2021spatial,gu2021exploiting}, designed networks \cite{dang2020detection, 2swe}, disentanglement learning \cite{liang2022exploring,yan2023ucf,yang2021learning}, reconstruction learning \cite{wang2021representative,cao2022end}, data augmentation \cite{li2020face,38_paper} and 3D decomposition \cite{Zhu_2021_CVPR}. Conversely, recent works in video forgery detection focus on optical flow \cite{Amerini_2019_ICCV}, eye blinking \cite{Li_2018_WIFS}, neuron behaviors \cite{Wang_2019_arXiv}, and temporal inconsistencies \cite{Haliassos_2021_CVPR, Wang_2023_CVPR}. 

\subsection{Open-set deepfake detection.}
As per the writer's knowledge, the only work done in open-set recognition for facial forgery detection is by Xu et al. \cite{xu2022supervised}. However, the method formulates the task as a binary classification problem (real vs. fake), without differentiating between multiple types of forgeries (e.g., fake$_1$, fake$_2$, etc.). As a result, it does not explicitly model the diversity of forgery distributions, which is critical for open-set recognition. Instead, the approach relies on learning shared representations across multiple forgery methods, thereby limiting its effectiveness. Furthermore, it employs feature fusion from multiple neural networks for representation learning, but does not investigate or incorporate dedicated open-set detection algorithms.

\subsection{Representation Learning}

Metric learning is a machine learning technique that represents images as points in a special space called ``latent space". In this space, each image is turned into a point, or ``embedding". The key idea is that the closer two points are in this space, the more similar their corresponding images are. If two images look alike, their points will be close together; if they are different, their points will be farther apart.

Using various methods, such as contrastive learning and supervised contrastive learning, we can define the important characteristics that the model should focus on. This allows the model to better understand and compare images based on these specific features.

In contrastive learning, we teach a model to recognize similarities and differences between images by creating ``similar" images through slight changes to the original, like rotating or cropping it \cite{chen2020simple}. These altered versions are treated as similar, while entirely different images are treated as dissimilar. This helps the model learn that certain objects remain similar even after transformations like rotation.

In supervised contrastive learning, labels are used to guide the model in understanding which images are similar and which are different \cite{supcon}. Images that belong to the same category (or class) are given embeddings (points in the latent space) that are close to each other, while images from different categories are assigned embeddings that are farther apart. This approach helps the model learn to group similar images and separate different ones based on their labels.

\section{Methods} \label{method}

Each deepfake image forgery is unique, depending on the generation method used, such as GANs or autoencoders. The main goal of the method is to design a latent space that captures only the forgery-specific characteristics of an image, ignoring irrelevant details such as identity, background, or lighting. By focusing solely on the forgery traits, the model avoids learning features that may be common across both real and manipulated data, thus improving its ability to distinguish forgeries from real images. This is achieved using weighted supervised contrastive learning, where forged images are assigned labels based on their specific forgery characteristics. The weighting ensures that the model prioritizes these forgery-specific features, leading to more cohesive and accurate representations. This approach helps the representation learning model to focus solely on distinguishing the most discriminative features relevant to deepfake detection.

\begin{figure}[htbp]
    \centering
    \begin{subfigure}[b]{1\linewidth}
        \centering
        \includegraphics[width=\linewidth]{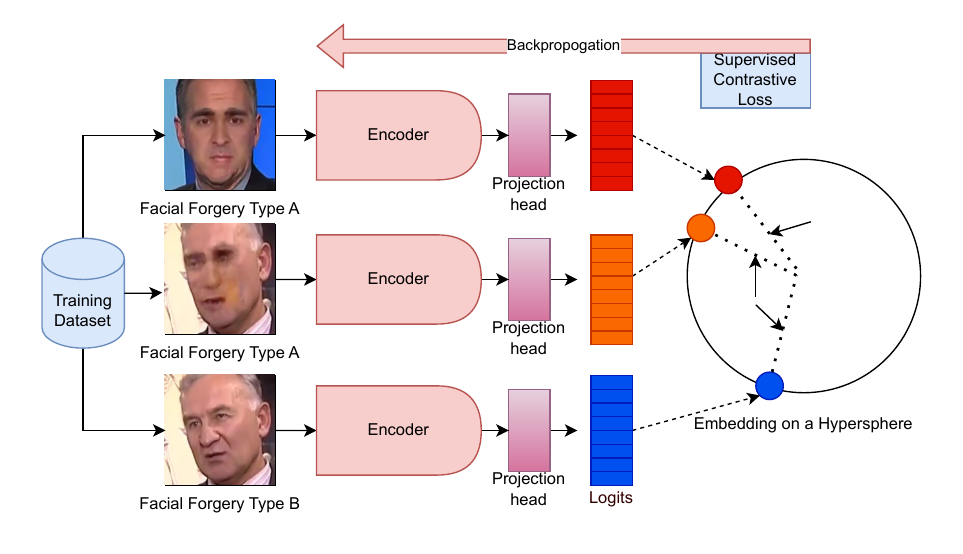}
        \caption{Stage 1: Supervised Contrastive Learning}
        \label{fig:sub1}
    \end{subfigure}
    \begin{subfigure}[b]{1\linewidth}
        \centering
        \includegraphics[width=\linewidth]{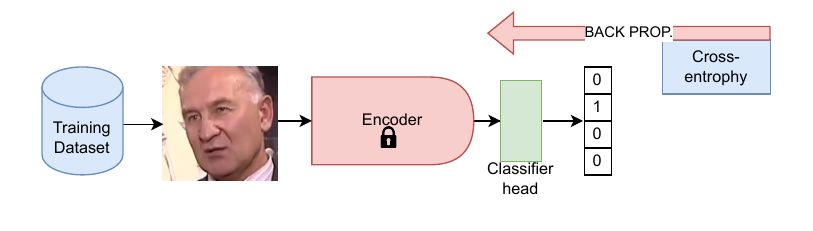}
        \caption{Stage 2: Supervised fine-tuning}
        \label{fig:sub2}
    \end{subfigure}
    \begin{subfigure}[b]{1\linewidth}
        \centering
        \includegraphics[width=\linewidth]{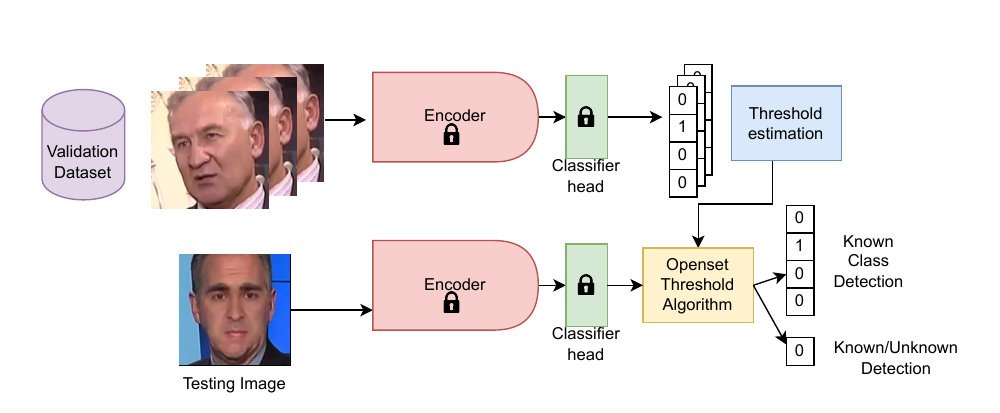}
        \caption{Stage 3: Open-set Threshold fine-tuning}
        \label{fig:sub3}
    \end{subfigure}
    \caption{ This figure shows the proposed methodology. (a) Supervised contrastive learning is used on the dataset to learn representations of various deepfake/real images via an encoder. (b) During the second stage, the encoder is frozen and a classifier is learnt on top of it. (c) In this phase, the logits of the training samples are employed to establish thresholds, which are then utilized for detecting unknown classes. }
    \label{fig:stacked}
\end{figure}

Once a latent space is constructed using weighted supervised contrastive learning in Stage 01, decision boundaries are established in Stage 02. These boundaries help address the open-set problem, where the model identifies unknown test classes that were not seen during training. In Stage 03, thresholds are determined based on confidence scores, allowing the model to detect whether a test sample belongs to one of the known classes or is an unknown class. This process is visualized as a three-stage pipeline in Fig.~\ref{fig:stacked}.

\subsection{Preliminaries/Supervised Contrastive Learning}

Let \(\{\boldsymbol{x}_k,\boldsymbol{y}_k\}_{k=1}^{N}\) represent \( N \) labeled samples from a given batch of training image data with C classes. From the given training batch, we create an augmented training batch of \( 2N \) samples (multi-viewed batch), represented as  \(\{\boldsymbol{\tilde{x}}_\ell,\boldsymbol{\tilde{y}}_\ell\}_{\ell=1}^{2N}\). 

In this multi-viewed batch,  \(\boldsymbol{\tilde{x}}_{2k-1}\) and \(\boldsymbol{\tilde{x}}_{2k}\) are two random augmentations of \(\boldsymbol{x}_k\) for each \(k=1, \ldots, N\).  These augmentations could include changes like rotations, flips, or color adjustments that preserve a subset of the original image's information. Both of these new images share the same label as the original image, meaning \(\boldsymbol{\tilde{y}}_{2k-1} = \boldsymbol{\tilde{y}}_{2k} = \boldsymbol{y}_k\) 

In addition to rotations, flips, or color adjustments, it implicitly treats different identities, poses, and lighting as augmentations of a particular type of specific deepfake.

The multi-viewed batch of images is first propagated through an encoder network, denoted as $Enc(\cdot)$, to obtain a 128-dimensional embedding $\boldsymbol{{r}_i} = Enc(\boldsymbol{{x}_i}) \in \mathcal{R}^{128}$. Subsequently, this embedding is processed through a projection network consisting of a linear layer, producing a vector $\boldsymbol{{z}_i} = Proj(\boldsymbol{{r}_i}) \in \mathcal{R}^{128}$ as illustrated in Fig.\ref{fig:stacked}. The supervised contrastive loss (\ref{eqn:supervised_loss_with_labels}) at the outputs of the projection network is used to train the model.

During our experiments, DenseNets gave the best results for the encoder choice \cite{DenseNet2017}.

\begin{equation}
  \mathcal{L}_{out}^{sup}(\tau,I, y) 
  = \sum_{i \in I} \frac{-1}{|P(i)|} \sum_{p \in P(i)} \log \left( \frac{\exp\left(\boldsymbol{z}_i \cdot \boldsymbol{z}_p / \tau\right)}{\sum\limits_{a \in A(i)} \exp\left(\boldsymbol{z}_i \cdot \boldsymbol{z}_a / \tau\right)} \right)
  \label{eqn:supervised_loss_with_labels}
\end{equation}

\textbf{Given:}

\begin{itemize}
    \item $I$ is the set of all samples in the batch.
    \item The index $i$ is the anchor.
    \item $A(i)$ is all the samples in the batch excluding the anchor. 
    \item $y_i$ is the label for sample $i$.
    \item The set of positive samples for sample $i$, denoted as $P(i)$, is:
    \[
    P(i) = \{ p \mid y_p = y_i \}
    \]
    This means that for every sample $p$ in the dataset, it belongs to $P(i)$ if its label $y_p$ matches the label of sample $i$, i.e., $y_p = y_i$.
    \item $\tau\in\mathcal{R}^+$ is a scalar temperature parameter
    
\end{itemize}

\subsection{Stage 1: Weighted Supervised Contrastive Learning} \label{stage01section}

In the context of supervised contrastive learning, the real data class poses unique challenges as it often fails to form compact representations in the feature space. To address this, we modify the supervised contrastive loss to emphasize the real class, encouraging tighter clustering of real data samples.

\begin{equation}
\mathcal{L}_{out}^{sup}(\tau, I, y) = \sum_{i \in I} \frac{-s(i,p)}{\sum_{a \in A(i)} s(i,a)} \sum_{p \in P(i)}   W(i,p,A(i))
\end{equation}

\begin{equation}
W({z}_i,{z}_p, A(i)) = \log \left( \frac{\exp\left(\boldsymbol{z}_i  \boldsymbol{z}_p / \tau\right)}{\sum_{a \in A(i)} \exp\left(\boldsymbol{z}_i \cdot \boldsymbol{z}_a / \tau\right)} \right)
\end{equation}

To further guide the representation learning, a pairwise similarity scaling factor \(s(i, j)\) is introduced to prioritize certain relationships between samples based on their manipulation status:

\begin{equation}
s(i,j) =
\begin{cases}
\alpha & \text{if } i \text{ and } j \text{ are both real}, \\
0 & \text{if } i \text{ and } j \text{ are not equal to each other} \\
1 & \text{otherwise.}
\end{cases}
\end{equation}

The contrastive loss is designed to place greater emphasis on clustering real data samples tightly in the feature space. By assigning higher weights \(s(i, j)\) to pairs of real data, the model learns more cohesive representations of real samples.

\subsection{Stage 2: Supervised fine-tuning}

For improved generalization and overall performance, the Stochastic Moving Average (SWA) encoder $\tilde{Enc}(\cdot)$ is constructed by averaging multiple trained encoder weights \cite{yang2019swalp}. 

Next, the input batch undergoes processing through the SWA encoder network $\tilde{Enc}(\cdot)$ to obtain embeddings. These embeddings are then fed into a classifier network $f(\cdot)$ and trained using cross-entropy loss. The classifier network predicts outputs $\boldsymbol{\hat{y}} = f(\boldsymbol{r}) \in \mathcal{R}^{D_P}$.

\subsection{Stage 3: The Open-Set Detection Algorithm}

In the third stage, the classifier $f(\cdot)$ learns to predict the probability of an input belonging to a specific class. This probability is denoted as $\hat{y_i}$, calculated by:
\begin{equation}
    y_i = P(y_i = 1 \mid x) = \frac{e^{f_i(\tilde{Enc}(x))}}{\sum_{j=1}^{K} e^{f_j(\tilde{Enc}(x))}} \label{eqn:ce_loss_with_labels}
\end{equation}

Here, $f_i(\cdot)$ represents the classifier's output for class $i$, and $y_i$ indicates whether the input's label belongs to class $i$.

The open-set algorithm determines rejection thresholds using a method called thresholding of the training dataset, detailed in Algorithm \ref{algo:thresholding}. This algorithm calculates thresholds for each class to determine whether a sample belongs to the class or should be rejected as unknown. It works by gathering confidence scores (softmax probabilities) of correctly classified training samples, then setting a threshold based on a specified percentile of these scores.

The process of open-set detection is explained in Algorithm \ref{algo:detection}. This algorithm checks if a test sample belongs to any known class or should be flagged as unknown. It compares the sample's confidence score for each class against the precomputed thresholds; if none are met, the sample is labeled as unknown.

This approach is advantageous because it does not require additional samples from new classes for fine-tuning, unlike traditional open-set detection methods.

\begin{algorithm}
\caption{Estimation of Class-wise Rejection Thresholds}
\begin{algorithmic}[1]

    \State Initialize empty set $T_i$ for each class $i$
    \For{each training example $(x, y)$}
        \State Calculate the softmax probabilities $\hat{y_i}$ using Eq.\ref{eqn:ce_loss_with_labels}
        \State Find the predicted class $i = \arg\max_i P(y_i = 1 | x) $
        \If{$y = i$}
            \State Add the output logit $\hat{y_i}$ to the set $T_i$
        \EndIf
    \EndFor
    \For{each class $i$}
        \State Sort the set $T_i$
        \State Calculate the $\lambda$ percentile of $T_i$ and record it as $\epsilon_i$
    \EndFor
\end{algorithmic} \label{algo:thresholding}
\end{algorithm}

\begin{algorithm}
\caption{Open-set classification}
\begin{algorithmic}[1]

    \For{each test sample $x$}
        \State Find the predicted logits $P(y_i = 1 | x) $ for all classes
        \State Set $is\_unknown \leftarrow True$
        \For{each class $i$}
            \If{$P(y_i = 1 | x)$$ \geq \epsilon_i$}
                \State $is\_unknown \leftarrow False$
                \State \textbf{break}
            \EndIf
        \EndFor
        \If{$is\_unknown$}
            \State Label $x$ as an unknown sample
        \EndIf
    \EndFor
\end{algorithmic} \label{algo:detection}
\end{algorithm}

\section{Experimental Evaluation} \label{experiomental}

This section details the data processing pipeline and evaluation strategies employed for both the Cross-manipulation and Cross-dataset evaluations, as presented in the results section.

\subsection{Datasets} \label{Datasets} 
\begin{figure}[ht]
        \centering
        \includegraphics[width=\linewidth]{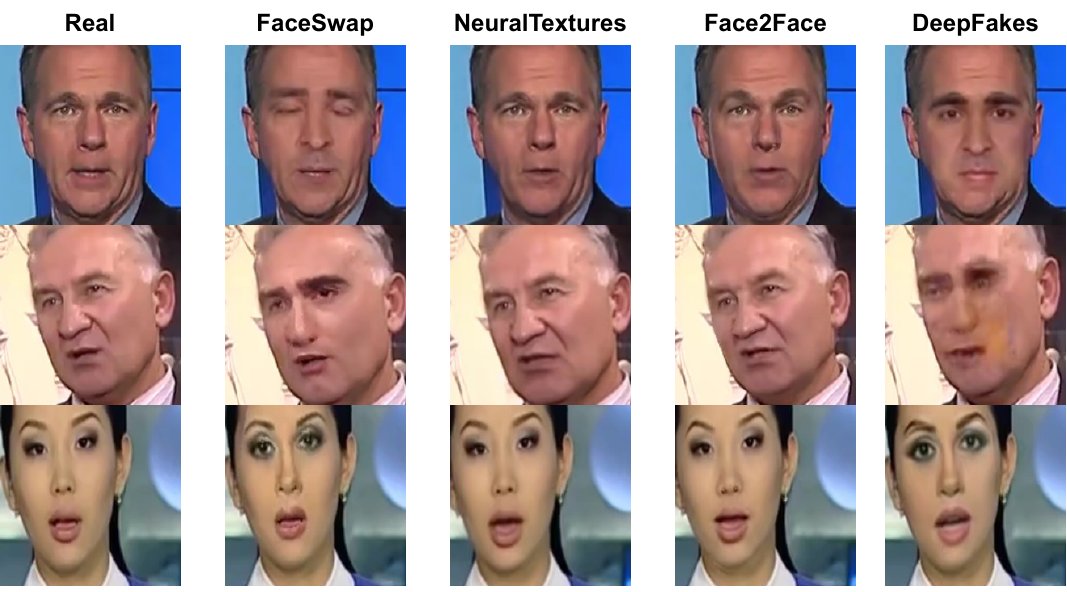}
        \label{fig:samplemeth}
    \caption{Sample face images from different facial forgery methods utilized in this study. The first column showcases pristine (non-manipulated) frames, while the subsequent four columns depict images generated by DeepFakes (DF), Face2Face (F2F), FaceSwap (FS), and Neural Textures (NT). DF and FS are identity swapping methods whereas NT and F2F are Facial Re-enactment methods.}
    \label{fig:both_images}
\end{figure}

The cross-manipulation evaluation utilizes the FaceForensics++23 (FF++c23) dataset \cite{Rossler2019FaceForensics}. FF++c23 consists of 1,000 pristine videos, each featuring a unique individual in a distinct background. Additionally, it includes 4,000 forged videos derived from these pristine videos, generated using four different face manipulation algorithms, as illustrated in Fig.~\ref{fig:both_images}: Face2Face (F2F) \cite{thies2016face2face}, DeepFakes (DF) \cite{deepfakes2020}, NeuralTextures (NT) \cite{thies2019deferred}, and FaceSwap (FS) \cite{kowalski2021faceswap}. DF (DeepFakes) and FS (FaceSwap) are identity-swapping methods that involve replacing the identity of a person in a video or image with that of another individual. In contrast, NT (NeuralTextures) and F2F (Face2Face) are facial reenactment methods, which manipulate the expressions or motions of a subject's face while preserving their original identity.

The cross-dataset evaluation incorporates both the FaceForensics++23 dataset and the CelebDF dataset \cite{Celeb_DF_cvpr20}.

\subsection{Baseline Model} \label{baseline}
Due to the novelty of our method, some results are compared against a baseline model. This baseline employs an Xception architecture and follows the exact training methodology outlined in \cite{Rossler2019FaceForensics}.

\subsection{Cross-manipulation Evaluation} \label{cross_man_section}

In real-world detection scenarios, defenders often lack detailed knowledge of the specific forgery techniques employed by attackers. Therefore, it is essential to evaluate the model's ability to recognize unfamiliar forgery methods as unknown. This assessment is particularly important in cross-manipulation settings, where the model encounters a variety of forgery techniques that were not present during training. Our approach aligns with the methodology proposed by Xu et al. \cite{xu2022supervised}.

The dataset has a total of five classes: the pristine videos and their four manipulated versions. For each class, its videos are split into three sets according to the protocol of Xu et al. \cite{xu2022supervised}: 600 for training, 200 for validation, and 200 for testing. Each pristine video and its respective four manipulated versions are kept together in the same set. Each video consists of a series of images called frames. Given that the algorithm processes images, we select 10 random frames from each video instead of using the entire video. We use multitask cascaded computational neural networks (MTCNN) \cite{zhang2016joint} to crop the face from the background. Afterwards, the cropped images are resized to a size of 224 × 224 pixels. Open-set classification aims to differentiate between known classes, which the model has been trained on, and unknown classes, which it has not encountered before. For our dataset, FF++c23, which has 5 classes, we evaluated four distinct combinations of known and unknown classes. Each combination is made by reserving one of the four facial forgery methods as unknown. For each combination, the model is trained on the training sets of the known classes and then tested on the test sets of the unknown and known classes. The performance metrics were then documented for each combination.

\subsection{Cross-Dataset Evaluation}

During the cross-dataset evaluation, the model is trained on all classes from the FF++c23 dataset. In this phase, the real class from the CelebDFv2 dataset is treated as the known "real" class, while the fake class is designated as the "unknown" class. This evaluation follows the same image processing methodology as the cross-manipulation evaluation.

\subsection{Metrics} \label{metrics_section}

We report the results as AUROC (Area Under the Receiver Operating Characteristic curve) and True Open-Set Classification (TOSC) Accuracy. 

\textbf{True-Open-Set-Classification (TOSC) Accuracy / Open-Set Classification Rate:} This accounts for the unknown samples. \begin{equation}
\text{TOSC} = \frac{\sum_{i=1}^{C+1} (TP_{i} + TN_{i})}{\sum_{i=1}^{C+1} (TP_{i} + TN_{i} + FP_{i} + FN_{i})} \quad 
\newline
\end{equation}
$TN_{i}$, $TP_{i}$, $FN_{i}$, $FP_{i}$ represent the True Negative, True Positive, False Negative, False Positive for a class $i$. The known classes are denoted by $\{1, 2, \ldots, C\}$. $C+1$ is the unknown test class.

\textbf{Area Under Receiver Operating Curve (AUROC).} To calculate the AUROC for open-set unknown classes, we evaluate how well a model can distinguish known classes (positive) from unknown classes (negative). This process requires adapting the traditional AUROC computation to the context of open-set recognition.  The AUROC for unknown classes is calculated following the methodology outlined in a recent survey \cite{Geng2020RecentAI}.

\section{Results} \label{results}

We examine deepfake classification across three open-set scenarios using the cross-manipulation evaluation described in Section~\ref{cross_man_section}: (A) detecting unknown deepfakes, (B) detecting known deepfakes, and (C) detecting deepfakes. The differences between these three cases are highlighted in Fig.~\ref{fig:mainidea}. The benchmark dataset used for comparison in this section is the widely recognized FaceForensics++c23 (FF++c23) dataset, which serves as the gold standard for deepfake detection research \cite{xu2022supervised}. 

Subsequently, we conduct additional experiments to assess cross-dataset performance using both the FaceForensics++c23 and CelebDFv2 datasets \cite{Rossler2019FaceForensics,Celeb_DF_cvpr20}.

\subsection{Open-set Classification of Unknown Deepfakes}

\begin{table}[htbp]
\centering
\tiny
\caption{Unknown class detection using the frame-level AUROC metric. The row is a model trained on the training classes and then tested on a test dataset containing an unknown class.  
 The proposed model significantly outperforms the Xception baseline in all the instances. Higher values of AUROC are optimal.}
\begin{tabular}{ l l c c }
\toprule
\textbf{Unknown Testing Class} & \textbf{Training Classes} & \textbf{Proposed} & \textbf{Xception} \\ 
\midrule
DF       &   F2F,FS,NT   &\textbf{ 0.7189  }          & 0.5184                 \\ 
F2F      &   DF,FS,NT  & \textbf{0.7399      }       & 0.5458                 \\ 
FS       &   DF,F2F,NT  & \textbf{0.5672  }          & 0.4671                 \\ 
NT       &   DF,F2F,FS  & \textbf{0.8233}              & 0.5985                 \\ 
\toprule
\end{tabular}

\label{tab:cross_man_auc_transposed}
\end{table}

\begin{table}[htbp]
\centering
\tiny
\caption{ Unknown class detection using the frame-level AUROC metric. The row is a model trained on the training classes and then tested on a test dataset containing an unknown class. Here the  
 The proposed model significantly outperforms the Xception baseline in all the instances. Higher values of AUROC are optimal. FE: Face reenactment, IS: Identity Swapping}
\begin{tabular}{l l l l c c} 
\toprule
\textbf{Unknown Testing Class} & \textbf{Type}  & \textbf{Training Classes}  & \textbf{Type} & \textbf{Proposed} & \textbf{Xception} \\ 
\midrule
FS   & IS    &   F2F,NT & FE & \textbf{0.4904 }          & 0.4781                 \\ 
DF   & IS  &   F2F,NT & FE & \textbf{0.8023}              & 0.5125                 \\ 
NT   & FE   &   DF,FS & IS & \textbf{0.7264}          & 0.5671                 \\ 
F2F  & FE     &   DF,FS  & IS & \textbf{0.7410}              & 0.5475                 \\ 
\bottomrule
\end{tabular}

\label{tab:cross_man_auc_transposed_crossent}
\end{table}

For each combination of training and testing classes based on FF++c23 (described in Section \ref{cross_man_section}), we measure the Area Under the ROC Curve (AUROC) for detecting the unknown test class, as shown in Table~\ref{tab:cross_man_auc_transposed}.

Our experimental results consistently demonstrate high AUROC values, indicating strong model performance in detecting unknown deepfake techniques. Furthermore, our method significantly outperforms the Xception baseline (detailed in Section~\ref{baseline}) across various training and testing class combinations. These findings highlight the robustness of our approach in open-set detection scenarios. For example, in Table \ref{tab:cross_man_auc_transposed}, when the unknown testing class is DF, the proposed model achieves an AUROC of 0.7189, compared to Xception’s 0.5184, representing a substantial improvement of over 20 percentage points. Similarly, for the NT unknown testing class, the proposed model attains an AUROC of 0.8233, significantly surpassing Xception’s 0.5985.

In Table~\ref{tab:cross_man_auc_transposed_crossent}, we further evaluate the method's performance in open-set scenarios where the unknown class belongs to a different type of deepfake method. These results confirm the robustness of our approach in detecting previously unseen types of deepfakes. For instance, in Table \ref{tab:cross_man_auc_transposed}, when the unknown testing class is DF—an identity-swapping method—and the training data contains face reenactment methods, the proposed model achieves an AUROC of 0.8023, compared to Xception’s 0.5125, marking a significant improvement of over 30 percentage points. Similarly, for the NT unknown testing class—a face reenactment method—the proposed model achieves an AUROC of 0.7264, vastly outperforming Xception’s 0.5671.

\subsection{Open-Set Classification of Known Deepfakes}

\begin{table}[htbp]
\centering
\tiny
\caption{This table shows the evaluation of known deepfakes with frame-level AUROC \% metric for our method.  We compare it with closed-set detectors referenced in \cite{Zhao2021LearningSF}, but the results are not strictly comparable due to the open-set nature of our method. For each method, the model's performance (Frame Level AUROC\%) is estimated on an equal split of the corresponding deepfake method and real data. Higher values are optimal.}

\begin{tabular}{ l l c c c c c c }
\toprule

\textbf{Methodology} & \textbf{Type of Model} & \textbf{Training Data} & \multicolumn{4}{c}{\textbf{Testing Classes}} \\ 
  &  & & \textbf{DF} & \textbf{F2F} & \textbf{FS} & \textbf{NT} \\ 
\midrule

MIL~\cite{Wang2018RevisitingMultiple} & Closed-set & FF++c23 & 99.51 & 98.59 & 94.86 & 97.96  \\ 
XN-avg~\cite{Rossler2019FaceForensics}  & Closed-set  & FF++c23 & 99.38 & 99.53 & 99.36 & 97.29  \\ 
Face X-ray~\cite{li2020face}  & Closed-set  & FF++c23 & 99.12 & 99.31 & 99.09 & 99.27  \\ 
S-MIL-T~\cite{Li2020SharpMultiple}  & Closed-set  & FF++c23 & 99.84 & 99.34 & 99.61 & 98.85  \\  
PCL + I2G~\cite{Zhao2021LearningSF}  & Closed-set  & FF++c23 & 100.00 & 99.57 & 100.00 & 99.58   \\  
\midrule
Proposed  & Open-set  & FF++c23, DF excluded  & -    & 97.87 & 99.43 & 93.07  \\ 
Proposed  & Open-set  & FF++c23, F2F excluded & 98.63 & - & 98.47 & 93.00 \\ 
Proposed  & Open-set   & FF++c23, FS excluded  & 99.15 & 98.64 & - & 95.1  \\ 
Proposed  & Open-set   & FF++c23, NT excluded  & 98.96 & 97.02 & 97.60 & -  \\ 
Proposed  & Open-set   & FF++c23  & 98.96 & 99.2 & 99.42 & 96.14  \\ 

\bottomrule 
\end{tabular}
\label{tab:inset_eval}
\end{table}
  
To evaluate our system, we analyze its ability to identify unseen test samples of known forgeries. This evaluation method, widely adopted in the literature, often emphasizes specialization over generalization.

The results presented in Table \ref{tab:inset_eval} show the evaluation of our proposed method on the FaceForensics++ dataset (FF++c23) across different manipulation types: DeepFake (DF), Face2Face (F2F), FaceSwap (FS), and NeuralTextures (NT). Our method demonstrates competitive results and performs comparably across various manipulations, particularly on FaceSwap (FS), where it achieves 0.9943 AUROC. For NeuralTextures (NT), the proposed method shows slightly lower performance (0.9512 to 0.9307) but remains competitive compared to other approaches.
Our method is an open-set classifier, whereas the others are closed-set classifiers. Despite this, our model still achieved competitive results, suggesting that the learned features are generalizable across manipulations.

\subsection{Open-set Classification of Deepfakes}
This focuses on detecting any deepfake (known or unknown) as a singular category. The TOSC (True Open-Set Classification from Section~\ref{metrics_section}) metric was employed, but instead of rejecting unknown instances as ``unknown", the model classified both known and unknown deepfakes as ``deepfake". This modification consolidates the task into detecting manipulation, regardless of whether it is known or unknown. It follows the protocol described in \cite{xu2022supervised}.

\begin{table}[htbp]
\centering
\tiny
\caption{TOSC Accuracy obtained on the unknown classes from various state-of-the-art approaches against proposed approaches in the cross-manipulation testing regime. Higher values are optimal. Results are taken directly from \cite{xu2022supervised}. The best results are highlighted in bold, and the second is underlined(N.R means that the original paper has not
reported the corresponding result).}
\begin{tabular}{ l c c c c }
\toprule
\textbf{Methodology} & \multicolumn{4}{c}{\textbf{Unknown Class}} \\ 
 & \textbf{DF} & \textbf{F2F} & \textbf{FS} & \textbf{NT} \\ 
 
\midrule
CLRNet \cite{tariq2020convolutional} & 50.12 & 53.73 & 50.00 & 69.75 \\ 
TAR \cite{tariq2021tar} & 75.25 & 72.90 & 51.65  & N.R \\ 
DDT \cite{aneja2020generalized} & 78.82 & N.R & N.R & 64.10 \\ 
Xception \cite{chollet2017xception}* & 82.73 & 64.69 & 49.74 & 55.59 \\ 
Fusion \cite{xu2022supervised} &\textbf{83.99} & 64.69 & 49.77 & 55.59 \\ 
Proposed & \underline{83.95} & \textbf{79.77}   & \textbf{76.12} & \textbf{81.22} \\ 
\bottomrule

\end{tabular}
\label{tab:TOSC_metric}
\end{table}

As seen in Table \ref{tab:TOSC_metric}, the proposed one-class model achieves state-of-the-art (SOTA) performance across several unknown classes. Compared to the baseline results reported in \cite{xu2022supervised}, our model consistently outperforms other methods. Notably, the proposed approach demonstrates superior accuracy in detecting both known and unknown deepfakes, achieving 83.95\% on the DF class and a substantial improvement across the other classes, with 79.77\% on F2F, 76.12\% on FS, and 81.22\% on NT. This places our model ahead of all compared methodologies in every scenario, delivering significant advancements in deepfake detection accuracy.

\subsection{Cross-dataset Performance and Evaluation.}
Most existing research has relied on the FF++c23 dataset, which serves as a suitable benchmark. However, it is crucial to evaluate models against other datasets for rigorous benchmarking. A key challenge in cross-dataset evaluation is the domain shift, particularly concerning real images. Real images from different datasets (such as FF++c23, CelebDF, or StarGAN) may vary in terms of lighting, resolution, compression artifacts, and even subject demographics. These differences can result in models trained on FF++c23 generalizing poorly when tested on other datasets, especially for real image classification. Furthermore, many datasets include a mixture of forgery methods, which introduces a blend of known and unknown manipulation techniques, making it difficult to track and evaluate the model's performance across these varying methods. These two implications mean it is difficult to evaluate the model's capacity as an unknown/deepfake known deepfake detector for datasets aside from FF++c23. Thus, we investigate its performance as a ``general deepfake detector". The results, as shown in Table~\ref{tab:table1crossdataset}, based on the CelebDF dataset, indicate that our method remains competitive \cite{Celeb_DF_cvpr20}, outperforming methods such as Meso4~\cite{afchar2018mesonet}, FAW~\cite{38_paper}, and Face X-ray~\cite{li2020face}.

\begin{table}
\centering
\tiny
\caption{Cross-dataset evaluations using the frame-level AUROC metric on the deepfake benchmark~\cite{yan2023deepfakebench}. All detectors are trained on FF++c23~\cite{Rossler2019FaceForensics} and evaluated on other datasets. }
\begin{tabular}{c c}
\toprule
\textbf{Detector} & \textbf{CDF-v2} \\
\midrule
Meso4~\cite{afchar2018mesonet} & 0.609 \\
MesoIncep~\cite{afchar2018mesonet} & 0.697 \\
CNN-Aug~\cite{he2016deep} & 0.703 \\
Xception~\cite{Rossler2019FaceForensics} & 0.737 \\
EfficientB4~\cite{tan2019efficientnet} & 0.749 \\
CapsuleNet~\cite{nguyen2019capsule} & 0.747 \\
FWA~\cite{38_paper} & 0.668 \\
Face X-ray~\cite{li2020face} & 0.679 \\
FFD~\cite{dang2020detection} & 0.744 \\
CORE~\cite{ni2022core} & 0.743 \\
Recce~\cite{cao2022end} & 0.732 \\
UCF~\cite{yan2023ucf} & 0.753 \\
F3Net~\cite{qian2020thinking} & 0.735 \\
SPSL~\cite{liu2021spatial} & 0.765 \\
SRM~\cite{luo2021generalizing} & 0.755 \\
Proposed & 0.695 \\
\bottomrule
\end{tabular}
\label{tab:table1crossdataset}
\end{table}

\section{Ablation Studies}

Extensive Ablation Studies are present in the Supplementary Material.  

\subsection{Impact of Representation Learning Method in Stage 1}

Several representation learning methods have been proposed in the literature. Cross-entropy (CE) remains the most widely used approach \cite{Mao2023CrossEntropy}. Recently, supervised contrastive learning (SupCon) has emerged as a more effective method for learning highly discriminative features, as demonstrated by its superior AUROC for unknown class detection and the well-structured organization of its latent space \cite{supcon}. Additionally, SIMCLR is a widely recognized unsupervised feature learning technique \cite{chen2020simple}.

The impact of these representation learning methods on detecting unknown classes is summarized in Table~\ref{tab:RepresentationTable}, where SupCon is shown to outperform the other approaches.

\begin{figure}[htbp]
    \centering
    \begin{subfigure}[b]{0.45\linewidth}
        \centering
        \includegraphics[width=\linewidth]{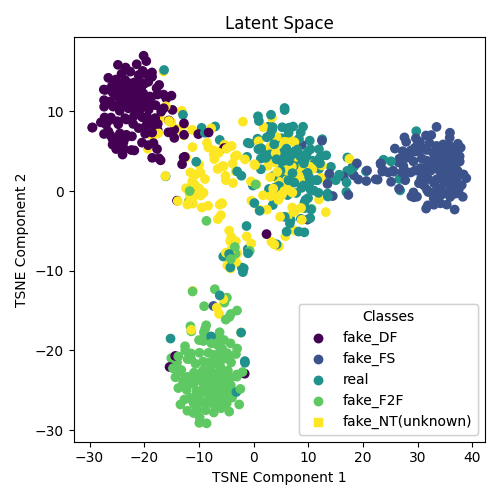}
        \caption{Weighted Supervised Contrastive Learning}
        \label{fig:supconviz}
    \end{subfigure}
    \hfill
    \begin{subfigure}[b]{0.45\linewidth}
        \centering
        \includegraphics[width=\linewidth]{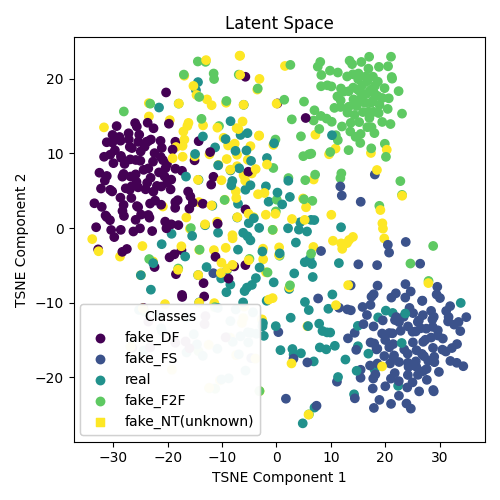}
        \caption{Cross Entropy}
        \label{fig:ceviz}
    \end{subfigure}
    \caption{TSNE visualizations of the learned representations for different training objectives.}
    \label{fig:stackedrep}
\end{figure}

Comparing the discriminative features learned by SupCon and CE is crucial for understanding their relative effectiveness. t-SNE visualizations illustrate that SupCon achieves superior separation of unknown features from real class features, demonstrating greater intra-class compactness and inter-class separation. The t-SNE plots for each method are shown in Fig.\ref{fig:supconviz}, and Fig.\ref{fig:ceviz}, with the unknown class NT and manipulated images highlighted in yellow. A major challenge for the encoder is avoiding the misclassification of unknown classes as unmanipulated (real). SupCon effectively addresses this issue, as evidenced in Fig.~\ref{fig:supconviz} ( In the figure, there is less mixing of the green(real) and yellow (unknown) points.), where unknown classes are clearly distinguished from both real and manipulated classes compared to cross entropy (Fig.~\ref{fig:ceviz}).  This robustness in feature learning underscores the documented strengths of supervised contrastive learning \cite{supcon}.

\begin{table}[htbp]
\centering
\tiny
\caption{Unknown class detection using the frame-level AUROC metric. Each column represents an unknown class, and each row shows the model’s AUROC performance on that class. Higher AUROC values are optimal.}
\begin{tabular}{ l c c c c }
\toprule
\textbf{Representation Learning Method} & \multicolumn{4}{c}{\textbf{Unknown Class}} \\ 
 & \textbf{DF} & \textbf{F2F} & \textbf{FS} & \textbf{NT} \\ 
 
\midrule
\textbf{SupCon} & 0.7009  & 0.7129 & 0.611 & 0.7877 \\ 
\textbf{Cross Entropy} & 0.6787 & 0.5888 & 0.5564 & 0.6313 \\ 
\textbf{SIMCLR} & 0.5012 & 0.5045 & 0.5034 & 0.4998 \\ 
\midrule
\textbf{Proposed} & \textbf{0.7189 } & \textbf{0.7399} & \textbf{0.678} & \textbf{0.8233} \\ 
\bottomrule
\end{tabular}
\label{tab:RepresentationTable}
\end{table}

\section{Conclusion and future work} \label{lmitations}

We propose an open-set deepfake detection framework based on supervised contrastive learning. Experiments using the publicly available FaceForensics++23 dataset demonstrate the effectiveness of our approach, achieving comparable or superior performance on detecting images forged by known and unknown deepfake techniques. Future work can address the following limitations: As shown by cross-dataset evaluations, the current method is sensitive to discrepancies between the real training and testing data distributions, particularly when the real data experiences domain shifts. Adopting a video level approach, incorporating more localized priors for facial features, could theoretically alleviate these issues observed during cross-dataset testing and should be explored in future research.

\section{Acknowledgment}

NB acknowledges Melbourne Graduate Research Scholar- ship.
The authors would like to thank Yu Xia, Jayanie Bogahawatte and Chathura Jayasankha for proof-reading the document. This research was supported by The University of Melbourne’s Research Computing Services .

{
    \small
    \bibliographystyle{ieeenat_fullname}
    \bibliography{main}
}


\end{document}